\documentclass[10pt,conference]{IEEEtran}
\IEEEoverridecommandlockouts
\usepackage{cite}
\usepackage{amsmath,amssymb,amsfonts}
\usepackage{algorithmic}
\usepackage{graphicx}
\usepackage{steinmetz}
\usepackage{textcomp}
\usepackage{mathtools}
\usepackage{threeparttable}
\usepackage{balance}
\usepackage{url}
\usepackage{hyperref}
\usepackage[table]{xcolor}
\newcolumntype{M}[1]{>{\centering\arraybackslash}m{#1}}
\newcommand{\mycc}{\cellcolor{lightgray}}
\newcolumntype{C}{>{\centering\arraybackslash}p{5cm}}
\def\BibTeX{{\rm B\kern-.05em{\sc i\kern-.025em b}\kern-.08em
    T\kern-.1667em\lower.7ex\hbox{E}\kern-.125emX}}
\begin{document}

\title{\textbf{FuNToM:} \textbf{Fu}nctional Modeling of RF Circuits Using a \textbf{N}eural Network Assisted \textbf{T}w\textbf{o}-Port Analysis \textbf{M}ethod}

\author{\IEEEauthorblockN{Morteza~Fayazi, Morteza~Tavakoli~Taba, Amirata~Tabatabavakili, Ehsan~Afshari, Ronald~Dreslinski}
Email: fayazi@umich.edu, tmorteza@umich.edu, vakili@umich.edu, afshari@umich.edu, rdreslin@umich.edu
\IEEEauthorblockA{\textit{University of Michigan, Ann Arbor, USA}}}
\maketitle
\begin{abstract}
Automatic synthesis of analog and Radio Frequency (RF) circuits is a trending approach that requires an efficient circuit modeling method. This is due to the expensive cost of running a large number of simulations at each synthesis cycle. Artificial intelligence methods are promising approaches for circuit modeling due to their speed and relative accuracy. However, existing approaches require a large amount of training data, which is still collected using simulation runs. In addition, such approaches collect a whole separate dataset for each circuit topology even if a single element is added or removed. These matters are only exacerbated by the need for post-layout modeling simulations, which take even longer. To alleviate these drawbacks, in this paper, we present FuNToM, a functional modeling method for RF circuits. FuNToM leverages the two-port analysis method for modeling multiple topologies using a single main dataset and multiple small datasets. It also leverages neural networks which have shown promising results in predicting the behavior of circuits. Our results show that for multiple RF circuits, in comparison to the state-of-the-art works, while maintaining the same accuracy, the required training data is reduced by 2.8x - 10.9x. In addition, FuNToM needs 176.8x - 188.6x less time for collecting the training set in post-layout modeling.
\end{abstract}


\section{Introduction}
The growing demand for analog and Radio Frequency (RF) circuits, due to their broad applications, has led to a crucial need for automated circuit synthesis~\cite{fayazi2021applications}. There are two classical automated synthesis approaches for analog and RF circuits, \textit{i.e.} simulation- and model-based, which both need a circuit modeling tool. In the simulation-based approaches, the circuit modeling tool is invoked multiple times to evaluate the circuit and generate new parameter candidates to meet the desired specifications~\cite{hassanpourghadi2021circuit}. In addition, the modeling tool is used to generate the training set of the model-based approaches. Since both methods invoke the modeling tool frequently (especially the simulation-based), using the common modeling tool, \textit{i.e.} SPICE simulation, is drastically time-consuming. Furthermore, technology scaling has caused much longer run times in both schematic and post-layout simulations~\cite{liu2021specification}.

Artificial Intelligence (AI) methods are an alternative to SPICE because of their speed and relative accuracy~\cite{liu2020transfer}. However, current approaches require a large amount of training data, which are still collected using simulation runs. So, they face the same time-consuming problem during training. Bayesian Model Fusion (BMF) is one of the common analog circuits functional modeling methods~\cite{li2012efficient,wang2013bayesian,fang2014bmf}. Neural Networks (NNs) also have shown promising results in all circuit design levels such as single-board computer, sizing, layout, System-on-Chip (SoC) design, and performance modeling~\cite{fayazi2022fascinet, fayazi2023angel, gusmao2020semi, ajayi2020open,fukuda2017op}. 

Despite all these advances in functional modeling of circuits, current works~\cite{li2012efficient,wang2013bayesian,fang2014bmf,daems2003simulation, mcconaghy2005caffeine, fukuda2017op,li2012large, alawieh2018efficient, hassanpourghadi2021circuit} train directly from circuit design parameters (capacitance value of capacitors, size of transistors, etc.) to Performance of Interests (PoI). Therefore, such conventional modeling approaches require a unique training set for each topology, because design parameters change if the topology changes. This means their whole training process needs to be redone even if a single element is added or removed from the circuit. Moreover, they assign an individual feature to each design parameter, which makes their models high-dimensional for modern complex analog and RF circuits. The matter is only exacerbated by the post-layout modeling, where collecting data takes longer.

To combat all the aforementioned challenges in circuit functional modeling, we propose an NN-assisted two-port analysis-based method, FuNToM. The goals of FuNToM are threefold: (a) decreasing the required number of new training sets after modifying a circuit topology; (b) further reducing the number of training sets for each topology; (c) achieving a fast, accurate RF functional modeling method.

\begin{figure}[t]
\centering
\includegraphics[width=\columnwidth]{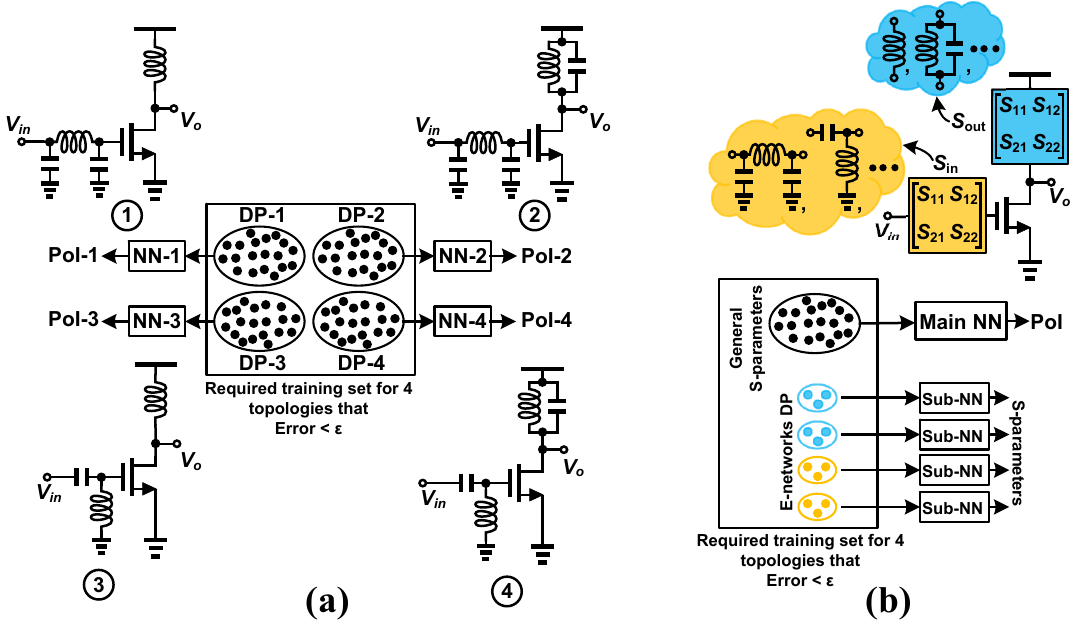}
\caption{An overview, concept, and the impact on the training set of (a) the conventional analog and RF circuits functional modeling method vs (b) the proposed modular S-parameters-based approach. The conventional method needs a whole separate dataset and model for each circuit topology while the proposed approach works properly for many circuit topologies with a single general format E-networks' S-parameters dataset.
\newline
$\begin{bmatrix}
S_{11} & S_{12}\\
S_{21} & S_{22}
\end{bmatrix}$: S-parameter matrix.
\newline
\newline
DP: Design parameters.}
\label{fig:general_approach}
\end{figure}

In order to decrease the required number of new training sets after modifying a circuit topology, we propose a method to properly model multiple topologies while using a single main dataset and multiple small datasets. For this purpose, we divide the circuit into multiple Electrical-networks (E-networks) and analyze them modularly via NNs. An E-network is a collection of electrical components, \textit{e.g.} capacitors, resistors, etc., that are interconnected~\cite{anderson2013network}. To modularly analyze E-networks, we leverage the Scattering parameters (S-parameters) concept. The S-parameter is a well-known powerful concept in RF circuits that is defined using two-port analysis and describes the circuit behavior of E-networks. In other words, in two-port analysis, the S-parameter is a $2\times2$ matrix that summarizes the circuit behavior of the associated E-network, such as input/output return losses, and insertion loss/gain~\cite{pozar2011microwave}. Therefore, the S-parameter can be replaced with the associated E-network in the circuit analysis.

The main advantage of modularly modeling the circuit is its generality. In our case, by passing the S-parameters to the NN, the NN learns the relationships between the general format of E-networks' S-parameters and the PoI. Since the S-parameter works as a wrapper around the E-networks, the NN accurately determines the functionality of the circuit with any E-networks' topology. In other words, training via S-parameters is a one-time process which works properly for many circuit topologies because of its modularity.

There are two types of NN models in our approach: 1) One main circuit NN for determining the PoI using S-parameters of E-networks; 2) multiple sub-NNs that each determines the S-parameters of the associated E-network using the circuit design parameters within such E-network. Creating new datasets for sub-NNs is the only required modification in the dataset of the proposed approach when the circuit topology changes as the training set of the main NN works for all topologies. However, since the circuit is divided into multiple E-networks and each E-network has a separate dataset, the size of such datasets is small because each deals with only a few design parameters. This dividing of the circuit is another advantage of the proposed method that is especially prevalent when going from schematic to the post-layout modeling. Instead of collecting a new large dataset by running time-consuming post-layout simulations for the whole circuit with many design parameters, it is only needed to gather small datasets for the E-networks, each with only a few design parameters. Fig.~\ref{fig:general_approach}(a) demonstrates the conventional modeling method using NNs, while Fig.~\ref{fig:general_approach}(b) shows the big picture of the proposed approach. As it is shown, our approach requires much less training data than the conventional method to have the same accuracy in modeling of four different topologies. Note that the circuit elements (design parameters) that are not part of E-networks, \textit{e.g.} the transistor in Fig.~\ref{fig:general_approach}(b), are also considered in our proposed model as explained with more details in Section~\ref{sec:nn_models}. FuNToM supports all RF circuits, including active and passive components, as multiple of them are evaluated in Section~\ref{sec:eval}.

We use the Circuit-Connectivity-Inspired-NN (CCI-NN) structure~\cite{hassanpourghadi2021circuit} in our main model to further reduce the number of training sets for each topology. As stated by Hassanpourghadi~\textit{et~al.}~\cite{hassanpourghadi2021circuit}, CCI-NN structure requires less training data in comparison with the Fully-Connected NNs (FC-NNs). Moreover, by condensing the design parameters using S-parameters, we may have a smaller design space than the conventional approach. If the number of design parameters in each E-network is more than the indexes in the S-parameter matrix, the NN model of the proposed approach will have fewer features than the conventional method and hence, less training sets will be required to have the same accuracy. This mainly has an impact on large circuits.

To validate our proposed method, FuNToM is tested by modeling the functionality of multiple phase shifters at the schematic level as well as multiple two-stage Low-Noise-Amplifiers (LNAs) at both schematic and post-layout levels. Our results show that using FuNToM, the required training set is reduced by a factor of 2.8x - 10.9x while maintaining the same accuracy in comparison to the state-of-the-art works. Also, the time for collecting the training set in the post-layout modeling using FuNToM is 176.8x - 188.6x faster. The results illustrate that FuNToM achieves an average R2-score of 0.95 when tested on 3,200 samples.

The main contributions of this paper can be summarized as follows:
\begin{itemize}
  \item Proposing FuNToM, an efficient RF functional modeling method which achieves an average R2-score of 0.95 when it is tested on 3,200 samples with different topologies and specifications.
  \item Reducing the number of required training set points compared to the state-of-the-art works by 2.8x - 10.9x by using two-port analysis and dividing the circuit into multiple E-networks.
  \item Decreasing the time for collecting the training set in the post-layout modeling by 176.8x - 188.6x compared to state-of-the-art works.
\end{itemize}

\section{Background \& Related Work}
\begin{figure}[t]
\centering
\includegraphics[width=\columnwidth]{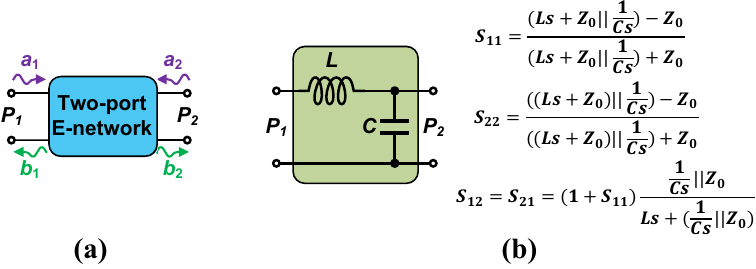}
\caption{(a) Excited and incident waves in a two-port E-network which are used for calculating S-parameters. (b) An example of a simple L-C network with the calculated S-parameters. s = j2$\pi$f where f is the frequency. $Z_0$ is the reference impedance (the impedance of excitation ports) which equals to 50 $\Omega$ in most cases.}
\label{fig:sparams}
\end{figure}

\subsection{Analog and RF Circuits Functional Modeling}
The goal in analog and RF circuits functional modeling is to approximate a PoI vector ($y$) of a given circuit by a function ($\hat{f}$) of the design parameters vector ($\mathbf{x}$), while the modeling error ($e$) is minimized. For this purpose, the SPICE is invoked as the ground truth function ($f$). In other words,
\begin{gather}
\label{eq:model_goal}
y = f(\mathbf{x}, t)\\
\left\{
\begin{array}{c}
y =\hat{f}(\mathbf{x}, t) + e, \\\nonumber
\text{subject to: } \text{minimize } e.\\
\end{array} \right.
\end{gather}
Note that $t$ is used to account for time/frequency variant PoIs. Power gain is an example of a PoI for an LNA, while resistance values of resistors are examples of the design parameters vector. The advantage of the functional modeling methods over SPICE is that once they get trained, they can determine the functionality of circuits, even when the design parameters change, in near zero time. This is because the circuit behavior is stored as weights for the functional modeling methods.

For minimizing the modeling error, different models, such as posynomials regression, genetic programming, BMF, and NN, have been studied~\cite{daems2003simulation, mcconaghy2005caffeine, li2012efficient, fukuda2017op}. The fusion in BMF means passing information (prior knowledge) from the schematic model to the post-layout model to reduce the expensive post-layout performance modeling cost~\cite{fayazi2021applications}. This passed information is then combined with a few post-layout training data points to solve the model coefficients via Bayesian inference~\cite{anzai2012pattern}.

The other goal of the functional modeling methods is to minimize the number of times that SPICE is invoked since it is time-consuming. Several studies have attempted to reduce the number of training samples while maintaining the same error, such as sparse regression, Co-Learning BMF (CL-BMF), hierarchical CL-BMF, and Circuit-Connectivity-Inspired NN (CCI-NN)~\cite{li2012large, alawieh2018efficient, hassanpourghadi2021circuit}. In hierarchical Co-Learning BMF, the circuit is partitioned into multiple stages and CL-BMF is applied to each stage~\cite{alawieh2018efficient}. In this regard, they consider lower dimensional models and as a result the training set size is decreased. Among all aforementioned works, NNs have shown the lowest modeling error however, they need a large training set~\cite{hassanpourghadi2021circuit}. Similar to~\cite{alawieh2018efficient}, Hassanpourghadi~\textit{et~al.}~\cite{hassanpourghadi2021circuit} break down a circuit into multiple stages, but they apply NN instead of CL-BMF for modeling. Zhao~\textit{et~al.}~\cite{zhao2021efficient} first, model the transistor using NNs. By leveraging such a model, they perform DC and AC modeling for the circuit. Despite of the accuracy and efficiency, there is no word about the post-layout modeling.

\subsection{Two-port Analysis \& S-parameters}
\label{sec:two-port_analysis}

\begin{figure}[t]
\centering
\includegraphics[width=\columnwidth]{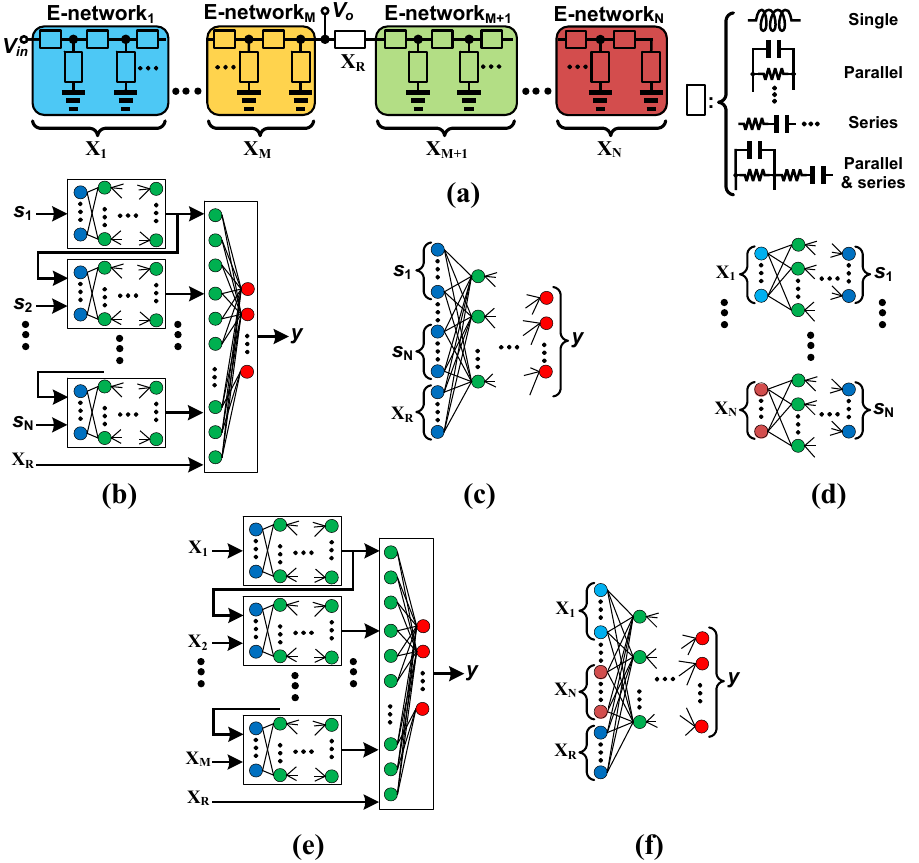}
\caption{(a) General circuit division into E-networks. Each rectangle is either a single element, parallel/series of elements, or a combination of parallel and series elements. (b) Main NN in the proposed approach for determining the PoI from S-parameters of E-networks (Equation~\eqref{eq:main_nn}) with CCI-NN structure. (c) Main NN in the proposed approach for determining the PoI from S-parameters of E-networks (Equation~\eqref{eq:main_nn}) with FC-NN structure. (d) Sub-NNs in the proposed approach for determining S-parameters from the design parameters of E-networks (Equation~\eqref{eq:sub_nn}). (e) The original CCI-NN model structure~\cite{hassanpourghadi2021circuit} with the circuit design parameters as inputs. (f) The conventional FC-NN model in the analog and RF functional modeling with the circuit design parameters as inputs.
\newline
$s_i$: S-parameters of the $i^{th}$ E-network. 
\newline
$\mathbf{x_i}$: Design parameters vector of the $i^{th}$ E-network. 
\newline
$\mathbf{x_R}$: A vector of the circuit's design parameters that are not included in any E-networks.}
\label{fig:nn_models}
\end{figure}

\begin{figure}[t]
\centering
\includegraphics[width=0.9\columnwidth]{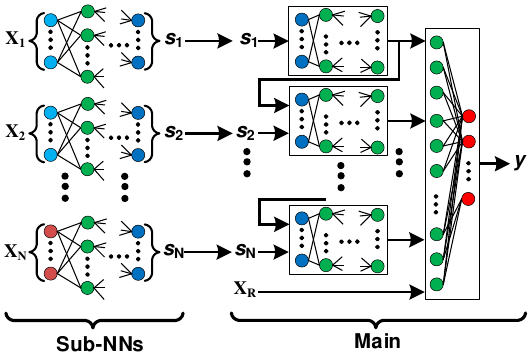}
\caption{Final FuNToM model by concatenating the sub-NNs and the main NN. 
\newline
$s_i$: S-parameters of the $i^{th}$ E-network. 
\newline
$\mathbf{x_i}$: Design parameters vector of the $i^{th}$ E-network. 
\newline
$\mathbf{x_R}$: A vector of the circuit's design parameters that are not included in any E-networks.}
\label{fig:final_nn}
\end{figure}

The two-port analysis is used to determine the response of a two-port E-network against the applied signals to its terminals. The circuit behavior of E-networks is determined using two-port analysis~\cite{pozar2011microwave}. Using two-port analysis, an S-parameter matrix is derived which is used for characterizing circuits. Fig.~\ref{fig:sparams} shows the procedure of determining S-parameters for a two-port E-network: each port is excited by incident waves, $a_1$ and $a_2$, and the reflected voltage waves, $b_1$ and $b_2$, are measured. The reflected voltage waves are a function of the incident waves:
\begin{equation}
\begin{aligned}
b_1& = S_{11}a_1 + S_{12}a_2, \\
b_2& = S_{21}a_1 + S_{22}a_2,
\label{eq:speq}
\end{aligned}
\end{equation}
where $S=\begin{bmatrix}
S_{11} & S_{12}\\
S_{21} & S_{22}
\end{bmatrix}$ is called the S-parameter matrix. 

S-parameters are calculated by setting $a_1$ and $a_2$ in Equation~\eqref{eq:speq} to zero one by one and getting the ratios of the reflected waves over the incident waves as follows:
\begin{equation}
\begin{aligned}
S_{11} & = \frac{b_1}{a_1}|_{a_2 = 0}, S_{12} = \frac{b_1}{a_2}|_{a_1 = 0}, \\
S_{21} & = \frac{b_2}{a_1}|_{a_2 = 0}, S_{22} = \frac{b_2}{a_2}|_{a_1 = 0}.
\end{aligned}
\end{equation}
It is noteworthy that setting $a_i = 0$ means that the $i^{th}$ port is terminated with a $Z_0$ impedance so that the reflection from that port ($a_i$) becomes zero. $Z_0$ equals 50 $\Omega$ in most cases. Fig.~\ref{fig:sparams}(b) demonstrates an example of S-parameters extraction of a simple L-C E-network. Based on the reciprocity theorem, $S_{12} = S_{21}$ for almost all passive E-networks such as the network mentioned in Fig.~\ref{fig:sparams}(b). Moreover, for the symmetric E-networks (see Fig.~\ref{fig:phase_shifter_example}(b)), $S_{11}=S_{22}$.

\section{Proposed Approach}
\subsection{Neural Network Models}
\label{sec:nn_models}
Breaking down the top-level circuit into multiple stages is used for simplifying $f$ in Equation~\eqref{eq:model_goal} and decreasing the number of training sets~\cite{hassanpourghadi2021circuit}. In our approach, we have gone further and we break down the circuit into multiple E-networks. There are two types of models in our approach: 1) One main model for determining the PoI using S-parameters of E-networks and design parameters that are not included in any E-networks; 2) multiple sub-models that each determines the S-parameters of the associated E-network using the circuit design parameters within such E-network. In other words:  
\begin{subequations}
\label{eq:nn}
\begin{align}
\label{eq:main_nn}
y & = \hat{g}([s_1, \ldots, s_N, \mathbf{x_R}], t) + e_1,\\
\label{eq:sub_nn}
s_i & = \hat{h_i}(\mathbf{x_i}, t) + e_{2,i} \quad i=1,\ldots,N,
\end{align}
\end{subequations}
where $s_i$ and $\mathbf{x_i}$ are the S-parameters and design parameters vector of the $i^{th}$ E-network ($s_i=[S_{11, i}, \ldots, S_{22, i}]$), respectively assuming there are $N$ E-networks. Indeed, $\mathbf{x_i}$ represents the value of circuit elements \textit{e.g.} capacitors, inductors, etc., inside the $i^{th}$ E-network. $\mathbf{x_R}$ is a vector of the circuit's design parameters that are not included in any E-networks \textit{e.g.} the transistor in Fig.~\ref{fig:general_approach}(b).

Fig.~\ref{fig:nn_models}(a) shows a general circuit division into multiple E-networks while Fig.~\ref{fig:nn_models}(b) and (c) illustrate two structures for our proposed main NN model. Both of them depict the NN representation of Equation~\eqref{eq:main_nn}. Fig.~\ref{fig:nn_models}(b) and Fig.~\ref{fig:nn_models}(c) have CCI-NN and FC-NN structures, respectively. As it is shown, the CCI-NN structure~\cite{hassanpourghadi2021circuit} breaks a giant NN into smaller chunks while considering the sequential paths and concatenating all at a final 1-layer NN. It should be mentioned that the CCI-NN is trained with the same dataset as the FC-NN \textit{i.e.} [($s_1, \ldots, s_N, \mathbf{x_R}$); $y$]. We analyze both FC-NN and CCI-NN structures for our main NN and select the one which gives the best results. As stated by~\cite{hassanpourghadi2021circuit}, CCI-NN requires less training data. Fig.~\ref{fig:nn_models}(d) depicts our sub-NN structures which are representations of Equation~\eqref{eq:sub_nn}.

\begin{figure}[t]
\centering
\includegraphics[width=\columnwidth]{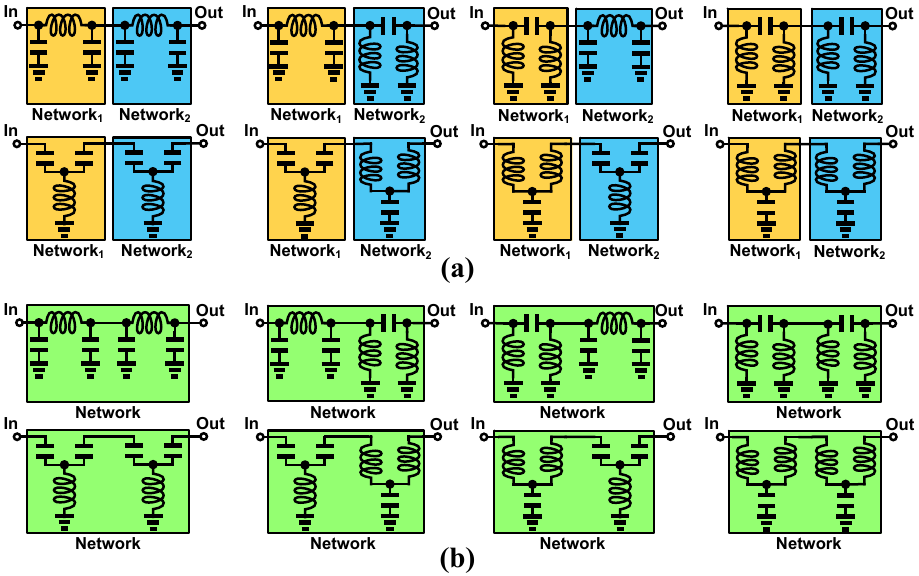}
\caption{Different ways for partitioning 8 topologies. (a) First scenario: each is divided into two E-networks. (b) Second scenario: there is only one E-network.}
\label{fig:partitioning}
\end{figure}

\begin {table}[t]
\begin{center}
\begin{threeparttable}
\centering
\caption{Required number of training data comparison between two scenarios of Fig.~\ref{fig:partitioning} for the main NN, sub-NNs, and total.}
\def\arraystretch{1.3}\tabcolsep 2pt
\label{table:partitioning_comparision}
\begin{tabular}{|M{40mm}|M{15mm}|M{15mm}|M{15mm}|}
\hline\hline
Scenario & Main NN & Sub-NNs & Total\\
\hline
First scenario (Fig.~\ref{fig:partitioning}(a)) & 1,800 & \mycc 400 & \mycc 2,200\\
\hline
Second scenario (Fig.~\ref{fig:partitioning}(b)) & \mycc 450 & 3,200 & 3,650\\
\hline
\end{tabular}
\end{threeparttable}
\end{center}
\end{table}

Our approach can be summarized as follows. The E-networks are identified and we assign a sub-NN to each. Then, the design parameters of each E-network are fed as inputs to the corresponding sub-NN for determining their S-parameters. In parallel, general S-parameters of E-networks are fed as an input to the main NN to determine the PoI. By combining these models (Equations~(\eqref{eq:main_nn}) and (\ref{eq:sub_nn}), \textit{i.e.} $\hat{g}([\hat{h_1}(\mathbf{x_1}, t),\ldots,\hat{h_N}(\mathbf{x_N}, t),\mathbf{x_R}], t)$), the PoI is determined based on the circuit design parameters. It should be noted that $\hat{g}$ in Equation~\eqref{eq:main_nn} is a function of $s_1,\ldots,s_N$ in general, not specifically for the $s_i$ in Equation~\eqref{eq:sub_nn}. In other words, Equation~\eqref{eq:main_nn} finds the relationship between the PoI and general S-parameters, and in parallel Equation~\eqref{eq:sub_nn} maps the design parameters to S-parameters in the desired circuit.

\begin{figure}[t]
\centering
\includegraphics[width=\columnwidth]{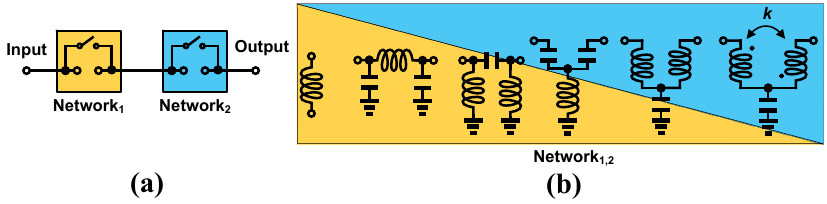}
\caption{Phase shifters tested on FuNToM. (a) General structure. (b) Network$_{1}$ and network$_{2}$ topologies.}
\label{fig:phase_shifter_example}
\end{figure}

\begin {table}[t]
\begin{center}
\begin{threeparttable}
\centering
\caption{Statistics of performance of interests for phase shifters.}
\def\arraystretch{1.3}\tabcolsep 2pt
\label{table:spec_stat_phase_shifter}
\begin{tabular}{|M{30mm}|M{13mm}|M{13mm}|M{12mm}|M{12mm}|}
\hline\hline
Specification & Min & Max & Average & SD\\
\hline
Input return loss [dB] & -117 & 0 & -4.87 & -10.12\\
\hline
Insertion loss [dB] & -133 & 0 & -3.17 & -10.26\\
\hline
Insertion phase [degree$^{\circ}$] & -180 & 180 & -23.02 & 84.5\\
\hline
Output return loss [dB] & -117 & 0 & -4.87 & -10.12\\
\hline
\end{tabular}
\end{threeparttable}
\end{center}
\end{table}

Equation~\eqref{eq:main_nn} is independent of the design parameters of E-networks. Therefore, it does not change if E-networks change, proving the generality and re-usability of our model for different E-networks. Furthermore, since each of the sub-NNs (Equation~\eqref{eq:sub_nn}) takes a subset of design parameters as inputs, they are simpler than the conventional NN model (Fig.~\ref{fig:nn_models}(e)). This significantly eases the collection of datasets for sub-NNs if E-networks change or during the post-layout modeling. Moreover, even in analyzing one specific circuit (without considering different E-networks), $\hat{g}$ may be simpler than $\hat{f}$ as S-parameters condense design parameters. This attribute is more pronounced in large RF circuits with many design parameters in each E-network. In addition, as mentioned in Section~\ref{sec:two-port_analysis}, $S_{12} = S_{21}$ for almost all passive E-networks and $S_{11}=S_{22}$ for the symmetric ones. These cause having even simpler $\hat{g}$. So, in such cases, the required training set is further reduced compared to the conventional method.

To summarize, in our proposed approach by concatenating the sub-NNs and the main NN, we determine the PoI by getting the circuit design parameters as inputs as shown in Fig.~\ref{fig:final_nn}. As illustrated, sub-NNs determine the S-parameters of the associated E-network by getting the circuit design parameters within such E-network as inputs while the main NN determines the PoI by getting S-parameters of E-networks as inputs. Two structures (CCI-NN in Fig.~\ref{fig:nn_models}(b) and FC-NN in Fig.~\ref{fig:nn_models}(c)) are analyzed for the main NN. As explained and our evaluations show, the CCI-NN structure gives the best results. The original CCI-NN structure proposed by~\cite{hassanpourghadi2021circuit} is shown in Fig.~\ref{fig:nn_models}(e) where the circuit design parameters are given as inputs and the PoI is the output. The conventional FC-NN model in the analog and RF functional modeling is also illustrated in Fig.~\ref{fig:nn_models}(f).

\subsection{Circuit Partitioning into E-networks}
There are multiple ways for partitioning a circuit into E-networks that each has some pros and cons. With decreasing the number of E-networks, the number of circuit elements at each increases as the total number of circuit elements is constant. So, the number of sub-NNs decreases while each gets more complicated. Usually, the required number of training data for a big complicated NN is more than the training data of multiple simple NNs. Moreover, with decreasing the number of E-networks, there would be fewer chunks in the CCI-NN structure of the main NN and since the input number of each chunk ($s_i=[S_{11, i}, \ldots, S_{22, i}]$) is almost fixed, the width of the main NN intuitively reduces. However, in order not to lose the accuracy, it may get deeper.

Similar to all machine learning techniques, to find the best results, different scenarios need to be tested. As an example to show different ways for partitioning a circuit, we consider two scenarios. To this end, we consider 8 different topologies. In the first scenario, each of these topologies is divided into two E-networks as shown in Fig.~\ref{fig:partitioning}(a). In the second scenario, the whole circuit is considered as one E-network as illustrated in Fig.~\ref{fig:partitioning}(b). Table~\ref{table:partitioning_comparision} compares the required number of training data between these two scenarios for the main NN, sub-NNs, and total to have the same accuracy. As it is summarized, the first scenario (Fig.~\ref{fig:partitioning}(a)) requires more data for the main NN while it needs less training data for sub-NNs and in total. Note that using such analysis and trying different scenarios, currently the circuit partitioning is done manually, but it can be replaced with an automated approach going forward.

\begin {table}[t]
\begin{center}
\begin{threeparttable}
\centering
\caption{Average R2-score of FuNToM models tested on different specifications of phase shifters shown in Fig.~\ref{fig:phase_shifter_example}.}
\def\arraystretch{1.2}\tabcolsep 2pt
\label{table:r2score_nn_phase_shifter}
\begin{tabular}{|M{11mm}|M{11mm}|M{14mm}|M{14mm}|M{14mm}|M{14mm}|}
\hline\hline
Model & Sub-NNs & \multicolumn{4}{|c|}{main NN}\\
\cline{3-6}
& & Input return loss & Insertion loss & Insertion phase & Output return loss\\
\hline
R2-score & 0.967 & 0.995 & 0.989 & 0.941 & 0.997\\
\hline
\end{tabular}
\end{threeparttable}
\end{center}
\end{table}

\begin{figure}[t]
\centering
\includegraphics[width=\columnwidth]{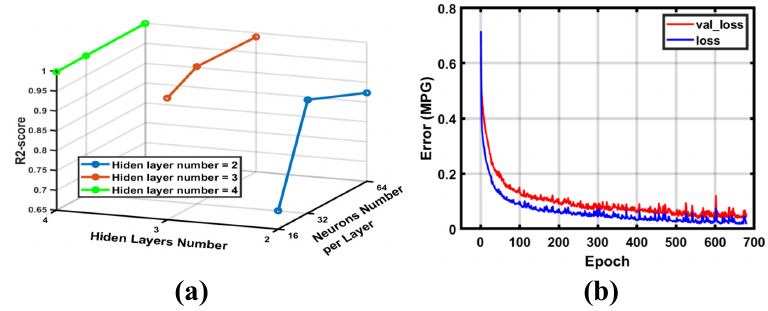}
\caption{Main model results for phase shifters. (a) R2-score comparison with different hyperparameters used for training. Three hidden layers with 32 neurons at each layer is the simplest model that gives a high R2-score of 0.99. (b) Loss vs. epoch diagram while three hidden layers with 32 neurons at each layer model is used for training.}
\label{fig:phase_shifter_results}
\end{figure}

\section{Evaluation}
\label{sec:eval}
In this section, we model the functionality of two different RF circuit types using FuNToM and compare the results with the state-of-the-art modeling methods. We model multiple phase shifters at the schematic level, and multiple two-stage LNAs at both schematic and post-layout levels. The design parameters include all transistor widths as well as capacitor and inductor values. The inductors and capacitors are used from PDK models of 55nm BiCMOS library, \textit{i.e.} quality factor and self-resonance of the inductors and capacitors are taken into account. All the SPICE simulations are run on a server with an NVIDIA TITAN V GPU and the frequency range for the simulations is 1 Hz-15 GHz. Also, all the NN models are built using the TensorFlow platform with the Adam optimizer and a learning rate of 0.001 and mean absolute error as the loss function. Moreover, all hidden layers have the RELU activation function. In order to avoid overfitting, the idea of early stopping~\cite{gulli2017deep} with the patience parameter of 125 is implemented. For this purpose, 10$\%$ of the data are used for validation during the training phase. In order to validate the results properly, a random separate test set with the size of 10$\%$ of the training set is used. As a default, all the results of FuNToM are based on the CCI-NN structure for the main NN unless that is mentioned.

\subsection{Phase Shifter}
Fig.~\ref{fig:phase_shifter_example}(a) depicts the general structure of the phase shifters tested on FuNToM. All combinations of E-networks shown in Fig.~\ref{fig:phase_shifter_example}(b) are used for both networks$_{1,2}$. Therefore, in total, FuNToM is trained and tested on 6$\times$6=36 different phase shifter topologies. There are on average 8 design parameters at each topology. The tested specification statistics are listed in Table~\ref{table:spec_stat_phase_shifter}. In total, 6,600 and 400 simulations are performed for the training set of the main NN and sub-NNs, respectively, per frequency. Since most of the specifications change with frequency, different datasets are needed to properly model the specifications over the frequency.

\begin{figure}[t]
\centering
\includegraphics[width=\columnwidth]{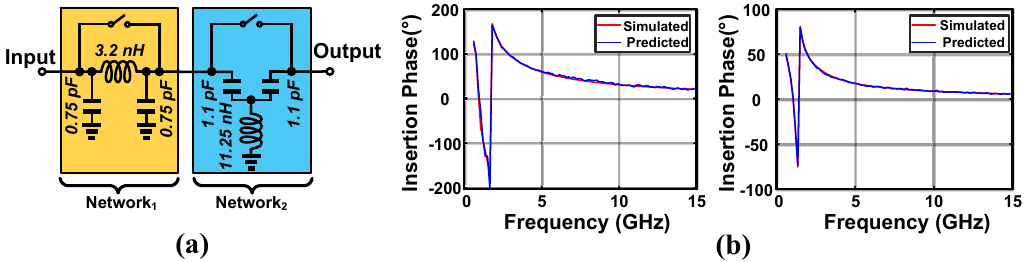}
\caption{An example of a phase shifter tested on FuNToM. (a) Topology. (b) The simulated (ground truth) and the predicted insertion phase by FuNToM over frequency. Left: when both switches are open; right: when both switches are closed.}
\label{fig:phase_shifter_test}
\vspace{-0.1in}
\end{figure}

\begin {table}[t]
\begin{center}
\begin{threeparttable}
\centering
\caption{Required number of training data, per frequency, comparison between the state-of-the-art works to achieve an average R2-score of 0.95. All models are tested on 1600 different phase shifters with the general structure of Fig.~\ref{fig:phase_shifter_example} in 36 topologies.}
\def\arraystretch{1.2}\tabcolsep 2pt
\label{table:modeling_comparision_phase_shifter}
\begin{tabular}{|M{20mm}|M{9mm}|M{9mm}|M{22mm}|M{22mm}|}
\hline\hline
Work & \cite{fukuda2017op} & \cite{hassanpourghadi2021circuit} & FuNToM FC-NN (Fig.~\ref{fig:nn_models}(c)) & \mycc FuNToM CCI-NN (Fig.~\ref{fig:nn_models}(b))\\
\hline
Training data number & 57,600 & 19,500 & 17,000 & \mycc 7,000\\
\hline
\end{tabular}
\end{threeparttable}
\end{center}
\end{table}

\begin{figure}[t]
\centering
\includegraphics[width=\columnwidth]{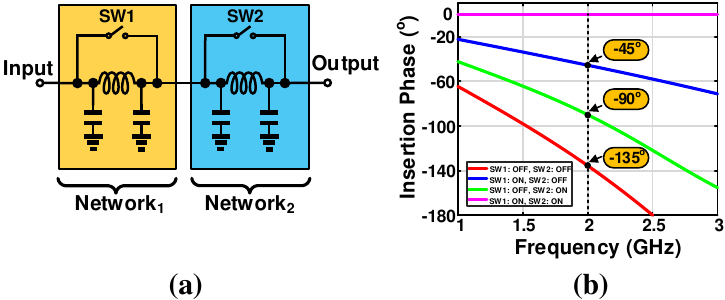}
\caption{Using FuNToM as the simulator in circuit sizing by NSGA-II algorithm to meet the desired specification in Table~\ref{table:optimization_phase_shifter}. (a) The given phase shifter topology. (b) Insertion phase results over frequency for different switch conditions.}
\label{fig:optimization_phase_shifter}
\end{figure}

\begin {table}[b]
\begin{center}
\begin{threeparttable}
\centering
\caption{An example of using FuNToM vs. SPICE as the circuit simulator for sizing Fig.~\ref{fig:optimization_phase_shifter}(a) circuit. Desired vs. generated specifications at 2 GHz frequency for different switch conditions are compared.}
\def\arraystretch{1.2}\tabcolsep 2pt
\label{table:optimization_phase_shifter}
\begin{tabular}{|M{9mm}|M{9mm}|M{32mm}|M{14mm}|M{14mm}|}
\hline\hline
SW1 & SW2 & Specification & \multicolumn{2}{|c|}{Simulator result}\\
\cline{4-5}
& & & FuNToM & SPICE\\
\hline
ON & OFF & Insertion phases = -45$^{\circ}$ & -45.61$^{\circ}$ & -45$^{\circ}$\\
\hline
OFF & ON & Insertion phases = -90$^{\circ}$ & -89.95$^{\circ}$ & -89.96$^{\circ}$\\
\hline
ON & OFF & Input return loss $<$ -35 dB & -43.5 dB & -70.6 dB\\
\hline
OFF & ON & Input return loss $<$ -35 dB & -37.8 dB & -61.3 dB\\
\hline
\multicolumn{3}{|c|}{Runtime} & 180s& 3,600s\\
\hline
\end{tabular}
\end{threeparttable}
\end{center}
\end{table}

The number of hidden layers and neurons per layer for the main model varies between 2-4 and 16-64, respectively. Fig.~\ref{fig:phase_shifter_results}(a) demonstrates a comparison between the R2-score achieved by different hyperparameters used for training the main model. Fig.~\ref{fig:phase_shifter_results}(b) also shows the loss vs. epoch of training and validation data of the main model. As it is shown, the training stops before overfitting happens. All sub-NNs have two hidden layers with 32 nodes at each layer.

\begin{figure}[t]
\centering
\includegraphics[width=\columnwidth]{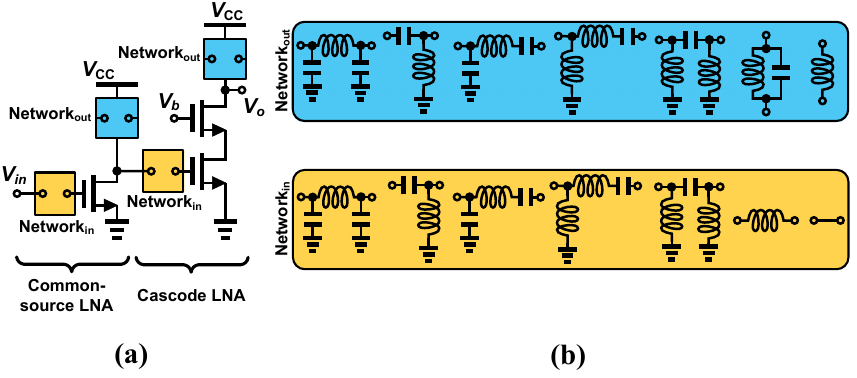}
\caption{Two-stage LNAs tested on FuNToM. (a) General structure; 1$^{st}$ stage is a common-source LNA while the 2$^{nd}$ stage is a cascode LNA. (b) Input and output matching networks for each stage.}
\label{fig:lna_example}
\end{figure}

Table~\ref{table:r2score_nn_phase_shifter} summarizes the average R2-score of sub-NNs and main NNs tested on different specifications of phase shifters. The average R2-score of all models is around 0.97. Moreover, Fig.~\ref{fig:phase_shifter_test}(a) depicts an example of a phase shifter tested on FuNToM. Fig.~\ref{fig:phase_shifter_test}(b) shows the predicted insertion phase as an example modeling specification by FuNToM over frequency compared to the simulated values.

The required number of training data per frequency is compared with the state-of-the-art works~\cite{fukuda2017op,hassanpourghadi2021circuit} while all achieve the same average R2-score of 0.95. We also consider both CCI-NN (Fig.~\ref{fig:nn_models}(b)) and FC-NN (Fig.~\ref{fig:nn_models}(c)) structures for the main NN of FuNToM. 1600 different phase shifters with the general structure of Fig.~\ref{fig:phase_shifter_example} in 36 topologies are given to all approaches for testing. Moreover, a wide range of design parameters is given to cover Table~\ref{table:spec_stat_phase_shifter}. As summarized in Table~\ref{table:modeling_comparision_phase_shifter},~\cite{fukuda2017op}, and~\cite{hassanpourghadi2021circuit} need 8.2x and 2.8x more training data than FuNToM, respectively. Moreover, FuNToM with FC-NN structure requires 2.4x more training data than FuNToM with CCI-NN structure.

In order to further demonstrate the functionality of FuNToM, we apply it as the circuit simulator to a circuit sizing method. For this purpose, we employ the well-known multi-objective genetic
algorithm NSGA-II~\cite{deb2002fast} to optimize the phase-shifter shown in Fig.~\ref{fig:optimization_phase_shifter}(a). An example of desired specifications at 2 GHz frequency is listed in Table~\ref{table:optimization_phase_shifter}. In the NSGA-II algorithm, the population size and number of generations are set to 30 and 30, respectively. The results when FuNToM and SPICE are used as the simulator in the sizing are summarized in Table~\ref{table:optimization_phase_shifter}. It should be mentioned that the FuNToM results are obtained when the generated circuit using FuNToM is verified by SPICE. FuNToM is 20x faster than SPICE, while both meet the desired specifications. The insertion phase over frequency of the sized circuit using FuNToM is shown in Fig.~\ref{fig:optimization_phase_shifter}(b) for different switch conditions. The network$_1$ and network$_1$ are designed to have an insertion phase of -45$^\circ$ and -90$^\circ$, respectively, at 2 GHz based on Table~\ref{table:optimization_phase_shifter}. As expected, the phase shifter insertion phase is $0^\circ$ when both switches are ON (meaning input and output are short-circuited). On the hand, when both switches are OFF, the insertion phase is -45$^\circ+$(-90$^\circ$) = -135$^\circ$. 

\subsection{Two-stage LNA}
\label{sec:eval_two-stage_LNA}
\begin {table}[t]
\begin{center}
\begin{threeparttable}
\centering
\caption{Statistics of performance of interests for LNAs.}
\def\arraystretch{1.2}\tabcolsep 2pt
\label{table:spec_stat_lna}
\begin{tabular}{|M{30mm}|M{12mm}|M{12mm}|M{12mm}|M{12mm}|}
\hline\hline
Specification & Min & Max & Average & SD\\
\hline
Transducer gain [dB] & -183.7 & 10.73 & -20.04 & 14.73\\
\hline
Power gain [dB] & -169.4 & 69.1 & -9.1 & 14.6\\
\hline
Available gain [dB] & -169.4 & 14.58 & -8.17 & 12.93\\
\hline
Noise Figure [dB] & 0.06 & 3082 & 17.15 & 18.48\\
\hline
Input return loss [dB] & -60 & 0 & -4.58 & -10.75\\
\hline
Output return loss [dB] & -60 & 0 & -4.58 & -11.06\\
\hline
Rollett's stability factor & -2.4$\times10^{5}$ & 10$\times10^{19}$ & 4.2$\times10^{12}$ & 4.5$\times10^{15}$\\
\hline
Stability factor & 2.1$\times10^{-4}$ & 2.01 & 0.36 & 0.33\\
\hline
Maximum power gain frequency [GHz] & 0.3 & 1.89 & 1.4 & 0.78\\
\hline
DC power [mW] & 2.4  & 2.7 & 2.54 & 0.15\\
\hline
\end{tabular}
\end{threeparttable}
\end{center}
\vspace{-0.1in}
\end{table}

\begin {table}[t]
\begin{center}
\begin{threeparttable}
\centering
\caption{Average R2-score of FuNToM models tested on different specifications of LNAs shown in Fig.~\ref{fig:lna_example}.}
\def\arraystretch{1.2}\tabcolsep 2pt
\label{table:r2score_nn_lna}
\begin{tabular}{|M{12mm}|M{11mm}|M{33.5mm}|M{28mm}|}
\hline\hline
Model & Sub-NNs & Common-source LNA main NN & Cascode LNA main NN \\
\hline
R2-score & 0.987 & 0.975 & 0.982\\
\hline
\end{tabular}
\end{threeparttable}
\end{center}
\vspace{-0.1in}
\end{table}

\begin{figure}[t]
\centering
\includegraphics[width=\columnwidth]{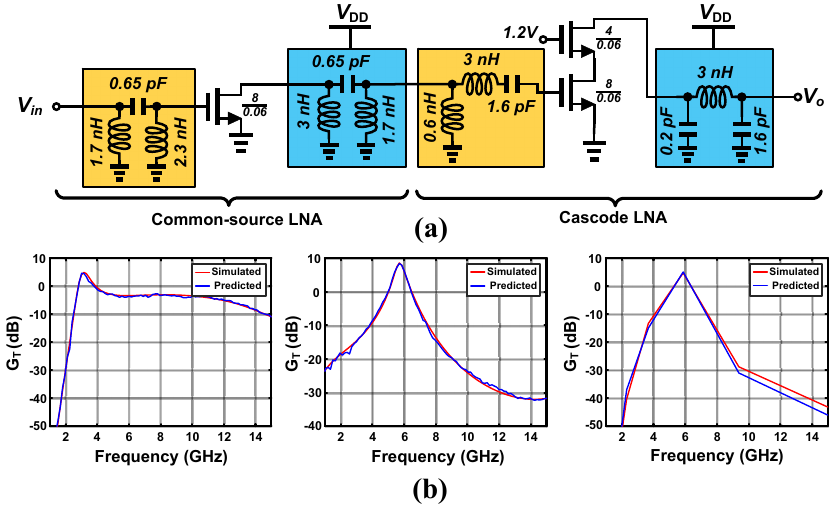}
\caption{An example of a two-stage LNA tested on FuNToM. (a) Topology; size of transistors are shown as $\frac{W(\mu)}{L(\mu)}$. (b) Schematic modeling: The simulated (ground truth) and the predicted transducer gain ($G_T$) by FuNToM over frequency. Left: Common-source (1$^{st}$) stage; middle: Cascode (2$^{nd}$) stage; right: the whole LNA (two-stage).}
\label{fig:lna_test}
\end{figure}

\begin{figure}[t]
\centering
\includegraphics[width=\columnwidth]{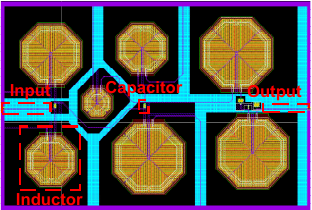}
\caption{The layout of Fig.~\ref{fig:lna_test}(a).}
\label{fig:lna_test_layout}
\end{figure}

\begin{figure}[t]
\centering
\includegraphics[width=\columnwidth]{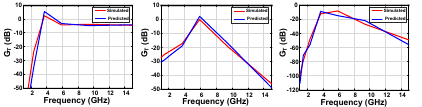}
\caption{Post-layout modeling of Fig.~\ref{fig:lna_test_layout}: The simulated (ground truth) and the predicted transducer gain ($G_T$) by FuNToM over frequency. Left: Common-source (1$^{st}$) stage; middle: Cascode (2$^{nd}$) stage; right: the whole LNA (two-stage).}
\label{fig:lna_test_pex}
\end{figure}

\begin {table}[t]
\begin{center}
\begin{threeparttable}
\centering
\caption{Required number of training data per frequency as well as the average time for collecting the training set in post-layout modeling per sample comparison between the state-of-the-art works to achieve an average R2-score of 0.95. All approaches are tested on 1600 different two-stage LNAs with the general structure of Fig.~\ref{fig:lna_example} that are selected from 2401 available topologies.}
\def\arraystretch{1.2}\tabcolsep 2pt
\label{table:modeling_comparision_lna}
\begin{tabular}{|M{34mm}|M{11mm}|M{11mm}|M{12mm}|M{12mm}|}
\hline\hline
Work & \cite{fukuda2017op} & \cite{hassanpourghadi2021circuit} & FuNToM FC-NN (Fig.~\ref{fig:nn_models}(c)) & FuNToM CCI-NN (Fig.~\ref{fig:nn_models}(b)) \mycc\\
\hline
Training data number & 7,200,000 & 160,000 & 48,100 & \mycc 14,600\\
\hline
Average time for collecting the training set in post-layout modeling per sample  & 5,790s & 5,430s & 30.7s & \mycc 30.7s\\
\hline
\end{tabular}
\end{threeparttable}
\end{center}
\vspace{-0.2in}
\end{table}

Fig.~\ref{fig:lna_example}(a) demonstrates the general structure of the two-stage LNAs tested on FuNToM. All combinations of E-networks shown in Fig.~\ref{fig:lna_example}(b) are used for input and output matching. So, in total, FuNToM is trained and tested on 7$\times$7=49 different LNA topologies at each stage which results in 49$\times$49=2401 total topologies. There are, on average, 14 design parameters at each topology. The tested specification statistics for the two-stage LNAs are summarized in Table~\ref{table:spec_stat_lna}. In total, 14,000 and 600 simulations are performed for the training set of the main NN and sub-NNs, respectively, per frequency.

The number of hidden layers and neurons per layer for the main model varies between 2-4 and 32-128, respectively. Our results show that three hidden layers with 32 neurons at each layer is the simplest model that gives a high R2-score of 0.97. Table~\ref{table:r2score_nn_lna} summarizes the average R2-score of sub-NNs and main NNs while tested on two-stage LNAs. Fig.~\ref{fig:lna_test}(a) shows an example of a two-stage LNA tested on FuNToM. Fig.~\ref{fig:lna_test}(b) illustrates the predicted transducer gain ($G_T$) as a sample modeling specification by FuNToM in the schematic modeling over frequency compared to the simulated values for each stage as well as the whole circuit. 

Fig.~\ref{fig:lna_test_layout} demonstrates the layout of Fig.~\ref{fig:lna_test}(a) which is used for post-layout modeling. As mentioned earlier, for the post-layout modeling we only need to gather training sets for sub-NNs, while the training set of the main NN is the same as what has been gathered for the schematic analysis. Moreover, the idea of transfer learning~\cite{liu2021specification} can be used to further reduce the number of training samples when going from schematic to the post-layout modeling. Fig.~\ref{fig:lna_test_pex} shows the predicted transducer gain ($G_T$) in the post-layout modeling by FuNToM over frequency compared to the simulated values for each stage as well as the whole circuit.

The required number of training data per frequency as well as the post-layout modeling training time of FuNToM are also compared with ~\cite{fukuda2017op,hassanpourghadi2021circuit} while all are achieving the same average R2-score of 0.95. 1600 different two-stage LNAs with the general structure of Fig.~\ref{fig:lna_example} that are selected from 2401 available topologies are given to all approaches for testing. Moreover, a wide range of design parameters is given to cover Table~\ref{table:spec_stat_lna}. As summarized in Table~\ref{table:modeling_comparision_lna},~\cite{fukuda2017op}, and~\cite{hassanpourghadi2021circuit} need 493.1x and 10.9x more training data than FuNToM, respectively. Moreover, FuNToM with FC-NN structure requires 3.3x more training data than FuNToM with CCI-NN structure. Furthermore, the average time for collecting the training set in the post-layout modeling of FuNToM is 188.6x and 176.8x less in comparison with~\cite{fukuda2017op}  and~\cite{hassanpourghadi2021circuit}, respectively. 

To calculate this average time, we measure the average post-layout simulation time for generating each training sample needed in each approach. In~\cite{fukuda2017op, hassanpourghadi2021circuit}, the post-layout simulation is run on the whole and half of the circuit, respectively, for generating each sample which is very time-consuming considering 4-8 inductors are included. However, in FuNToM, 14,000 of the required data are for the main NN which their post-layout simulations are very fast as they include only transistors and modular S-parameters. Moreover, the 600 training data for sub-NNs are gathered by simulating the E-networks that include only one or two inductors which are relatively fast. So, as the results show, the average time for running each simulation for FuNToM is much less than the other approaches. We save around 240,000 hours (160,000~$\times$~5,430s~$-$~14,600~$\times$~30.7s~=~240,000h) by using FuNToM instead of~\cite{fukuda2017op} in such a post-layout training.

\section{Conclusion}
This work presents a novel RF functional modeling method, FuNToM. FuNToM divides circuits into multiple E-networks and analyzes them modularly via NNs. Leveraging such a modular analysis using the concept of S-parameters has made the NN models used in FuNToM general enough that work for multiple circuit topologies. To validate our work, FuNToM is tested on more than 1,600 phase shifters and 1,600 two-stage LNAs. The results indicate that FuNToM reduces the size of the required training data by 2.8x - 10.9x in comparison with the state-of-the-art works while maintaining the same accuracy. Moreover, FuNToM needs 176.8x - 188.6x less time for collecting the training set in post-layout modeling.


\begin{thebibliography}{10}
\providecommand{\url}[1]{#1}
\csname url@samestyle\endcsname
\providecommand{\newblock}{\relax}
\providecommand{\bibinfo}[2]{#2}
\providecommand{\BIBentrySTDinterwordspacing}{\spaceskip=0pt\relax}
\providecommand{\BIBentryALTinterwordstretchfactor}{4}
\providecommand{\BIBentryALTinterwordspacing}{\spaceskip=\fontdimen2\font plus
\BIBentryALTinterwordstretchfactor\fontdimen3\font minus
  \fontdimen4\font\relax}
\providecommand{\BIBforeignlanguage}[2]{{%
\expandafter\ifx\csname l@#1\endcsname\relax
\typeout{** WARNING: IEEEtran.bst: No hyphenation pattern has been}%
\typeout{** loaded for the language `#1'. Using the pattern for}%
\typeout{** the default language instead.}%
\else
\language=\csname l@#1\endcsname
\fi
#2}}
\providecommand{\BIBdecl}{\relax}
\BIBdecl

\bibitem{fayazi2021applications}
M.~Fayazi \emph{et~al.}, ``Applications of Artificial Intelligence on the Modeling and Optimization for Analog and Mixed-Signal Circuits: A review,'' \emph{IEEE TCAS-I: Regular Papers}, 2021.

\bibitem{hassanpourghadi2021circuit}
M.~Hassanpourghadi \emph{et~al.}, ``Circuit Connectivity Inspired Neural Network for Analog Mixed-Signal Functional Modeling,'' in \emph{DAC}, IEEE, 2021.

\bibitem{liu2021specification}
J.~Liu \emph{et~al.}, ``From Specification to Silicon: Towards Analog/Mixed-Signal Design Automation using Surrogate NN Models with Transfer Learning,'' in \emph{ICCAD}, IEEE, 2021, pp. 1--9.

\bibitem{liu2020transfer}
J.~Liu \emph{et~al.}, ``Transfer Learning with Bayesian Optimization-Aided Sampling for Efficient AMS Circuit Modeling,'' in \emph{ICCAD}, IEEE, 2020.

\bibitem{li2012efficient}
X.~Li \emph{et~al.}, ``Efficient Parametric Yield Estimation of Analog/Mixed-Signal Circuits via Bayesian Model Fusion,'' in \emph{ICCAD}, IEEE, 2012.

\bibitem{wang2013bayesian}
F.~Wang \emph{et~al.}, ``Bayesian Model Fusion: Large-Scale Performance Modeling of Analog and Mixed-Signal Circuits by Reusing Early-Stage Data,'' in \emph{DAC}, IEEE, 2013, pp. 1--6.

\bibitem{fang2014bmf}
C.~Fang \emph{et~al.}, ``BMF-BD: Bayesian Model Fusion on Bernoulli Distribution for Efficient Yield Estimation of Integrated Circuits,'' in \emph{DAC}, IEEE, 2014, pp. 1--6.

\bibitem{fayazi2022fascinet}
M.~Fayazi \emph{et~al.}, ``FASCINET: A Fully Automated Single-Board Computer Generator Using Neural Networks,'' \emph{IEEE TCAD}, 2022.

\bibitem{fayazi2023angel}
M.~Fayazi \emph{et~al.}, ``AnGeL: Fully-Automated Analog Circuit Generator Using a Neural Network Assisted Semi-supervised Learning Approach,'' \emph{IEEE TCAS-I: Regular Papers}, 2023.

\bibitem{wang2018application}
Z.~Wang, X.~Luo, and Z.~Gong, ``Application of Deep Learning in Analog Circuit Sizing,'' in \emph{ICCSAI}, 2018, pp. 571--575.

\bibitem{gusmao2020semi}
A.~Gusm{\~a}o \emph{et~al.}, ``Semi-Supervised Artificial Neural Networks towards Analog IC Placement Recommender,'' in \emph{ISCAS}, IEEE, 2020.

\bibitem{ajayi2020open}
T.~Ajayi \emph{et~al.}, ``An Open-source Framework for Autonomous SoC Design with Analog Block Generation,'' in \emph{VLSI-SOC}, IEEE, 2020.

\bibitem{fukuda2017op}
M.~Fukuda \emph{et~al.}, ``OP-AMP sizing by inference of element values using machine learning,'' in \emph{ISPACS}, IEEE, 2017, pp. 622--627.

\bibitem{daems2003simulation}
W.~Daems \emph{et~al.}, ``Simulation-Based Generation of Posynomial Performance Models for the Sizing of Analog Integrated Circuits,'' \emph{IEEE~TCAD}, vol.~22,
  no.~5, pp. 517--534, 2003.

\bibitem{mcconaghy2005caffeine}
T.~McConaghy, T.~Eeckelaert, and G.~Gielen, ``CAFFEINE: Template-Free Symbolic Model Generation of Analog Circuits via Canonical Form Functions and Genetic Programming,'' in \emph{DATE}, IEEE, 2005.

\bibitem{li2012large}
X.~Li, W.~Zhang, and F.~Wang, ``Large-Scale Statistical Performance Modeling of Analog and Mixed-Signal Circuits,'' in \emph{CICC}, IEEE, 2012.

\bibitem{alawieh2018efficient}
M.~B. Alawieh \emph{et~al.}, ``Efficient Hierarchical Performance Modeling for Analog and Mixed-Signal Circuits via Bayesian Co-Learning,'' \emph{IEEE~TCAD}, vol.~37, no.~12, pp. 2986--2998, 2018.

\bibitem{anderson2013network}
B.~D. Anderson and S.~Vongpanitlerd, ``Network Analysis and Synthesis: A Modern Systems Theory Approach,'' \emph{Courier Corporation}, 2013.

\bibitem{pozar2011microwave}
D.~M. Pozar, ``Microwave Engineering," \emph{John wiley \& sons}, 2011.

\bibitem{anzai2012pattern}
Y.~Anzai, ``Pattern Recognition and Machine Learning,'' Elsevier, 2012.

\bibitem{zhao2021efficient}
Z.~Zhao and L.~Zhang, ``Efficient Performance Modeling for Automated CMOS Analog Circuit Synthesis,'' \emph{IEEE TVLSI}, vol.~29, no.~11, pp. 1824--1837, 2021.

\bibitem{gulli2017deep}
A.~Gulli \emph{et~al.}, ``Deep Learning with Keras,'' Packt Publishing Ltd, 2017.

\bibitem{deb2002fast}
K.~Deb, A.~Pratap, S.~Agarwal, and T.~Meyarivan, ``A Fast and Elitist Multiobjective Genetic Algorithm: NSGA-II,'' \emph{IEEE TEVC}, vol.~6, no.~2, pp. 182--197, 2002.

\end{thebibliography}

\ifCLASSOPTIONcaptionsoff
  \newpage
\fi
\bibliographystyle{IEEEtran}

\end{document}